# Tetris: Re-architecting Convolutional Neural Network Computation for Machine Learning Accelerators

Hang Lu[1,2], Xin Wei[2], Ning Lin[2], Guihai Yan[1,2] and Xiaowei Li[1,2]
State Key Laboratory of Computer Architecture, Institute of Computing Technology, Chinese Academy of Sciences[1]
University of Chinese Academy of Sciences[2]
{luhang, weixin, linning, yan, lxw}@ict.ac.cn

***Abstract***—Inference efficiency is the predominant consideration in designing deep learning accelerators. Previous work mainly focuses on skipping zero values to deal with remarkable ineffectual computation, while zero bits in non-zero values, as another major source of ineffectual computation, is often ignored. The reason lies on the difficulty of extracting essential bits during operating multiply-and-accumulate (MAC) in the processing element. Based on the fact that zero bits occupy as high as 68.9% fraction in the overall weights of modern deep convolutional neural network models, this paper firstly proposes a weight kneading technique that could eliminate ineffectual computation caused by either zero value weights or zero bits in non-zero weights, simultaneously. Besides, a split-and-accumulate (SAC) computing pattern in replacement of conventional MAC, as well as the corresponding hardware accelerator design called Tetris are proposed to support weight kneading at the hardware level. Experimental results prove that Tetris could speed up inference up to 1.50x, and improve power efficiency up to 5.33x compared with the state-of-the-art baselines.

## I. Introduction

Deep convolutional neural networks has driven significant progress in machine learning applications such as real-time image recognition and detection, neural language processing, etc. In order to bolster the accuracy, state-of-the-art deep convolutional neural network (DCNN) architectures embrace more complex connections and ever increasing number of neurons and synapses to deal with complicated classification tasks at a higher speed.

Given the limitations of the conventional general purpose architectures, many researchers propose specialized accelerators targeting specific computation patterns in modern DCNNs [1, 2, 3]. DCNN is comprised of multiple consecutively connected layers, from tens to even hundreds, and in each layer, the input feature activations and weights perform convolutions for each channel of each filter, which takes nearly 98% computations in the overall DCNN, accompanied by non-linear activations such as ReLu and pooling. As the major power and performance dominator, improving the computational efficiency of convolutions without compromising the robustness of the learning model is a critical step to enable efficient inference, especially on lightweight devices with limited resources and power budget like smartphones and autonomous robotics.

Conventionally, a plethora of techniques are proposed to utilize the irrelevance of each weight-activation pair, seeking to mine the potentials of multiply-and-accumulate (MACs) operations that could be executed in parallel, for as many as possible to attain an optimal computation throughput of the accelerator. However, due to the inherit nature of DCNNs, inference efficiency is susceptible to substantial *ineffectual computations* [4, 5], which lies in two aspects: firstly, the zero operands, in both weights and activations generated in the previous layer, are accepted as input for MAC operations of the current layer. These zero values are multiplied and added together with other non-zero operands, wasting time and energy but contribute nothing to the final output feature map. To address this issue, some approaches leverage the sparsity of the input data by intentionally skipping zero values or pruning near-zero values in the processing element (PE) [4, 6]. Although these value based optimization could be directly employed to alleviate ineffectual computation at a per-layer basis, "zero valued bits", as another form factor of the ineffectual computation, also occupies a large fraction of the input data set, and its consequence to the inference efficiency is not easy to be mitigated. The reason is that the structural and functional design of DCNNs relies on multiplication and takes it as one portion to form the final partial sum. Following this computing paradigm, modern accelerators allocate plenty of multipliers in PEs to maximize the throughput. Zero values could be conveniently ignored at the input of the multiplier; zero-valued bits however cannot be directly skipped over in performing multiplications in the multipliers. According to our exploration in Section II, non-essential bits (or 0s) in the input data set contributes as high as 68.9% ineffectual computation, which further exacerbates this problem.

To address this challenge, some prior approaches propose to use bit-level serialization when performing MACs [5, 7], taking advantage of the feature that fixed point multiplication could be broken down into a series of shifts and additions of single-bit multiplications. However, the essential bits may emerge at different locations in fixed-point presented weights or activations, so such scheme must rely on large shifters with varied length to cover the worst-case position of bit "1s", i.e. the $16^{th}$ bit in fix point 16 (fp16) weights. The shift operation lies on the critical path of the subsequent summing operation. Different weight or activation scenarios may yield different latencies in shifting to the target position, and hence yield unpredictable cycles to calculate a final summation. The hardware design is bound to cover the worst-case scenario that not only increases the design complexity but also suppresses the frequency potentials of the accelerator.

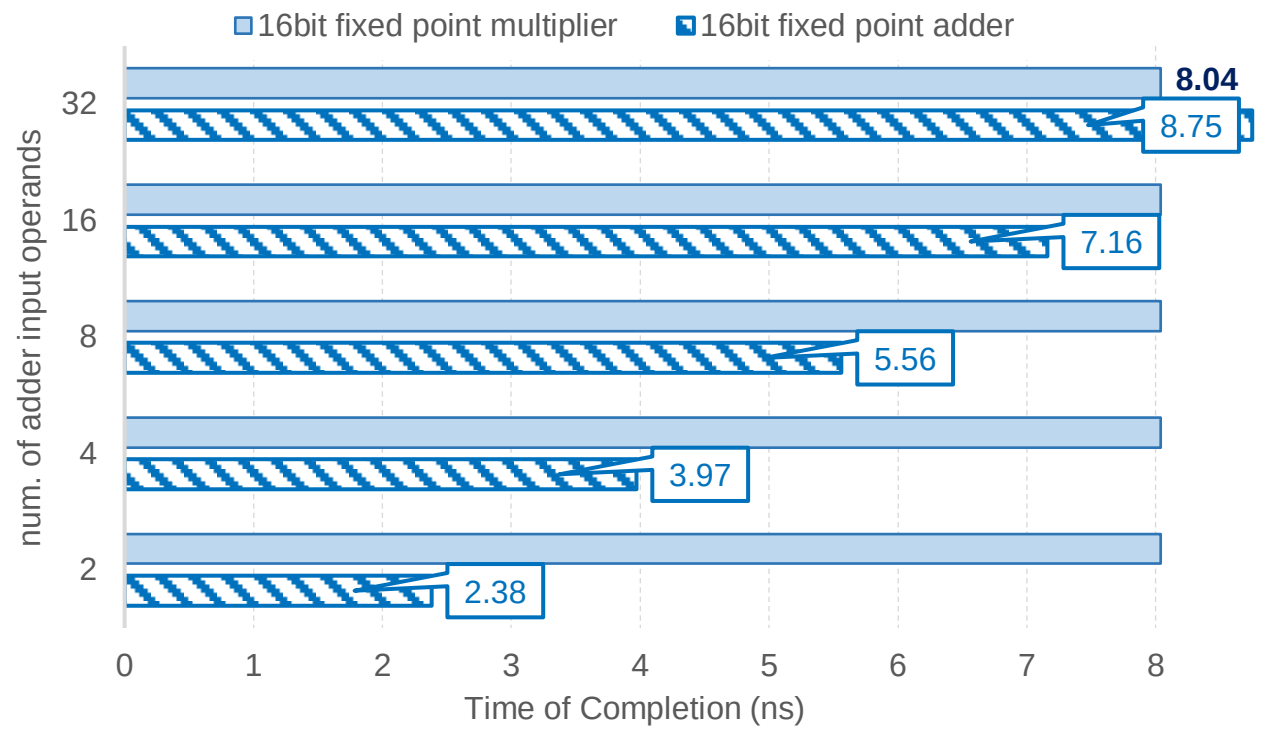

Figure 1 Temporal overhead of 16-bit fixed point adder w/ varied input operands, versus 16-bit fixed point multiplier with only 2 operands. The data is obtained from RTL simulation of Xilinx Z7020 FPGA using Vivado HLS tool.

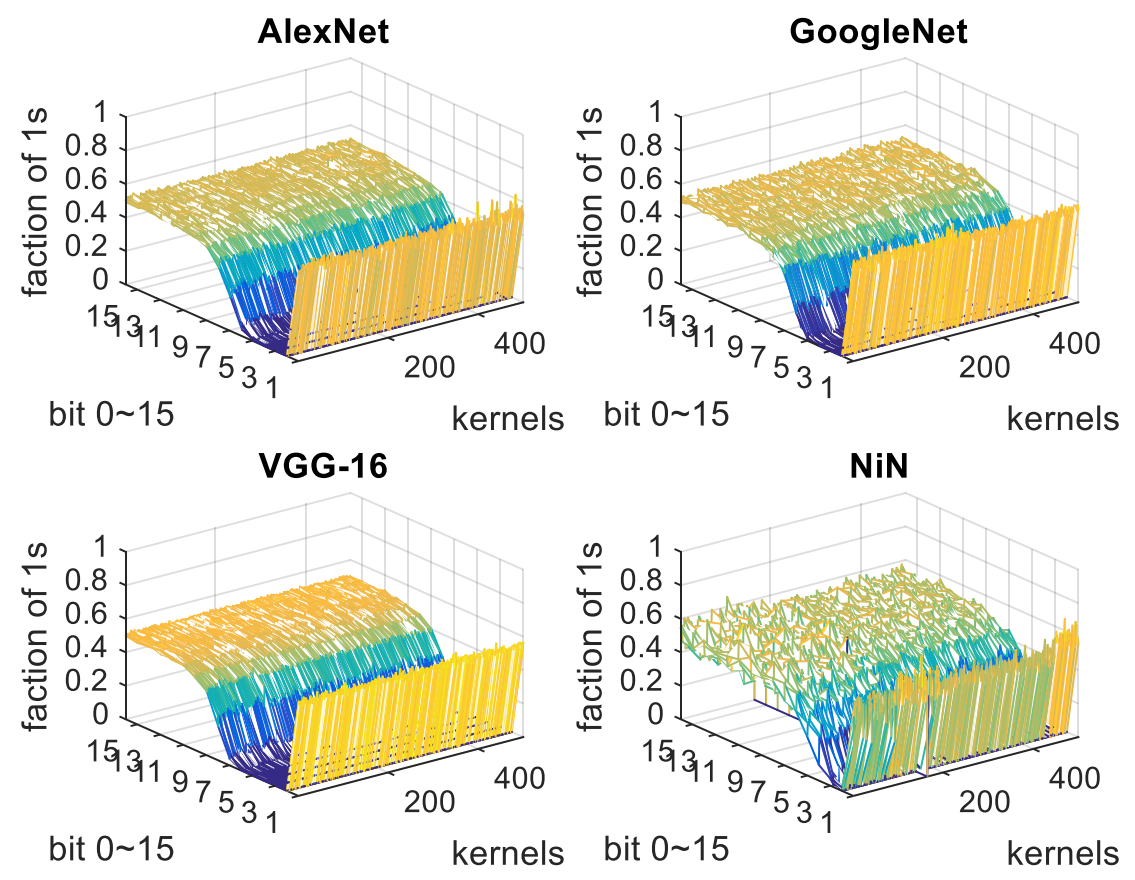

Figure 2 Essential bit (1s) distribution, across bit 0 ~ 15 for fp16 weights extracted from 500 kernels of 4 DCNN models.

Table 1 fraction of zero-valued weights & zero bits in all weights

| **Models** | **Zero Weights (%)** | **Zero BITs in Weights (%)** |
|---|---|---|
| AlexNet | 0.093 | 70.52 |
| GoogleNet | 0.050 | 65.23 |
| VGG-16 | 0.156 | 70.52 |
| VGG-19 | 0.182 | 71.09 |
| NiN | 0.193 | 67.02 |
| **GeoMean** | 0.135 | 68.88 |

Ineffectual computation, especially caused by zero-valued bits, limits the headroom of inference efficiency in terms of both computation throughput and power consumption. It calls for a new computing paradigm as well as the supportive hardware accelerator to circumvent the issues faced by the previous methods. In this work, we propose a high throughput deep learning accelerator design, called *Tetris*, targeting the effective computation during the inference of modern DCNN models. Tetris differs from previous MAC-based CNN accelerators by re-architecting the computing pattern of the convolutions. It abandons the preconceived design philosophy that pair-wise multiplication must be performed in order to obtain the exact intermediate partial sum, while oblivious to the ineffectual computation caused by the zero bits. Tetris firstly kneads a batch of weights in the lane eliminating the slack of zero bits, and only leaving essential bits. Subsequent partial sum computation is transformed to segment adding without multiplications or tedious shifting to the desired position, so the inference efficiency is significantly boosted. In general, this paper makes the following contributions:

- We propose a novel *weight kneading* technique for the effectual computation in machine learning accelerators. Weight kneading removes the zero bit slacks that prevail in synaptic weights in the modern DCNN models. Unlike data compression or pruning, it reduces the number of weights with no accuracy loss, and the inference cycles are also reduced.
- We re-architect the inference *computing pattern* of the DCNN models. Concretely speaking, we propose *Split-and-Accumulate* (SAC) in replacement of the de facto computing pattern – MAC. Multipliers are no longer involved but substituted by a series of low-cost segment adders. Shifting and summation that exists in traditional multipliers are performed only once after a batch of weight/activation pairs finish SAC, obtaining fruitful power and performance benefits.
- At last, we implement *Tetris* accelerator to mine the maximum potential of the weight kneading technique and SAC. Tetris organizes a series of SAC units, and consumes the kneaded weights and activations for the high throughput yet low power inference computing. Experiment via high level synthesis tool proves the efficacy of Tetris, compared with the canonical state-of-the-arts.

## II. Do we need MACs exactly?

### A. *Ineffectual Computations*

Convolution calculus is all about performing multiply-and-accumulates (MACs). In order to accelerate this kind of operation, classic DNN accelerators are architected following MAC computing pattern by deploying multipliers and adders at each neuron and synapsis lane. The multiplications could be either for floating point 32 operands or, in most modern accelerator designs [8, 9], 16-bit fixed point and int8 quantization to acquire a balanced inference efficiency and accuracy. Compared to the fixed point adder, multiplier dominates the latency of MACs. For a 2-operand multiplying, it takes 12.3% more time over the adder with even 16 adding operands as demonstrated in Figure 1. The latency stems from the shifting of synapsis (or weights) iteratively from the LSB of the activation till the MSB during multiplication, and worse still, different DNN models have various precision requirements at even a per-layer basis, so the multiplier designed for MAC must be able to cover the worst case latency, even for most of the time, the shifting and summation of intermediate values are not always contributive to the final result, a.k.a *ineffectual computations*.

As the major problem of MAC, ineffectual computations could be manifested in two aspects: the operands are *zero values* or including large portion of *zero bits*. Compared with zero bits, zero values occupy a trivial slice of input weights, as shown in Table 1. These small portion of 0s can be easily eliminated through advanced micro-architectural design, or compression techniques at the memory level and are avoided as the input of the multipliers. However, as the inherited nature of the fixed point multiplication, shifting to obtain each intermediate segment is agnostic of the "zero-valued bits", another major factor of the ineffectual computation. Table 1 shows that compared with the essential bits (or 1s), the fraction of zero bits is as high as 68.9% on average, which means avoiding the impacts of zero bits could have potentials in boosting the inference efficiency in both performance and power, significantly.

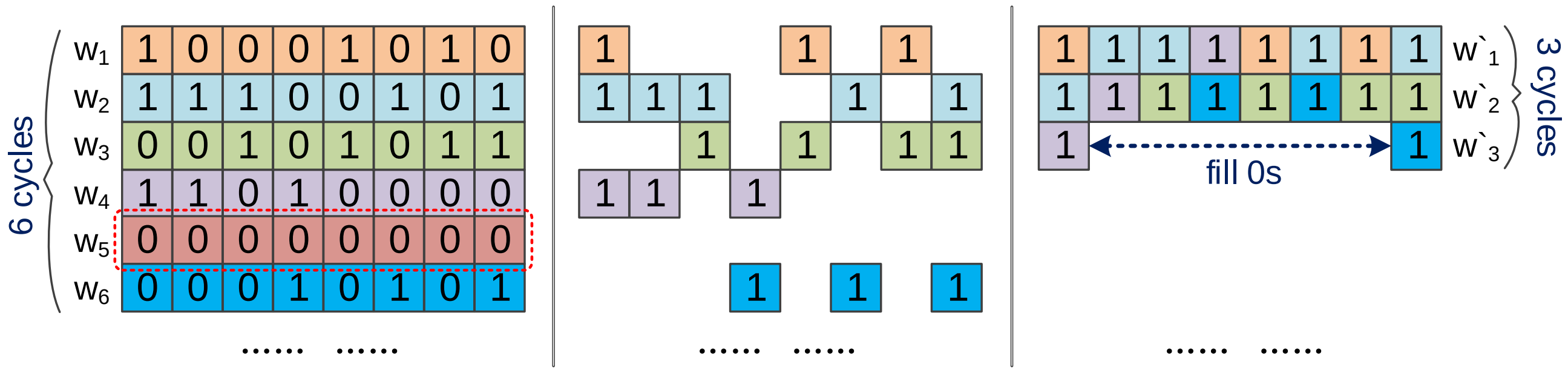

Figure 3 Weight Kneading. The design philosophy lies on squeezing the slacks in consecutive weights. (a) shows the scenario of input weights w/ zero slacks before kneading; eliminating slacks and only essential bits are shown in (b); weights after kneading is shown in (c).

Although there are different implementations that leverage quantization techniques by transforming the weights into power of two values [10], or more aggressive ternary [11], or even binary values [12, 13, 14] and multiplications are hence downgraded to pure shifting or adding operations, these schemes suffer from severe accuracy loss at large datasets. High precision MACs are sometimes necessary and how to effectively optimize zero values and zero-valued bits simultaneously remains as a critical problem in DNN accelerator design.

### *B. Harnessing Slacks*

Conventional CNN accelerators seek to process a series of weights and activations in the lane in batches by allocating certain amount of PEs capable of absorbing as large as 256 weight/activation pairs in total [9]. Each pair is feed into its PE and finishes MAC within one cycle, with ineffectual computation of zero bits also accounted. If we could make use of the time allocated for ineffectual computation but use it for essential computation, it would be definitely beneficial for the PE throughput. In this paper, we term the multiplication of 0s in PE as "slacks".

In Figure 2, we evaluate the proportion of essential bits across different bit locations in the weights of 4 commonly used DCNN models. We select 500 kernels across different layers and found that the distribution of 1s demonstrates two similar behaviors: (1) the portion of essential bits remains nearly identical at each bit location, around 50% ~ 60%, which also means 40% ~ 50% portion are slacks at these locations. No bit location exhibits radically large portions of essential bits; (2) certain bits exhibit large portion of slacks. For example, bit 3 ~ 5 only have less than 1% essential bits. The "cliff" at these locations proves that some bits are almost comprised of slacks but multiplier does not differentiate them with essential bits when performing MAC. If we want an augmented inference efficiency, the slacks must be harnessed.

The streamlined distribution of essential bits or slacks provides a unique opportunity to squeeze the slacks at a per-bit level. In specific, if the slacks presented in previous weight could be replaced by essential bits (1s) of the subsequent weight, we could replace the ineffectual computations with essential contributive computations and process multiple pairs in one cycle. Figure 2 has proved the headroom for squeezing these weights could be as high as 50%, and it does not emerge any roofline at any bit location so the overall weights could be compressed into nearly half of their initial volume. In other words, we could save 50% time during inference. However, it would be tricky to achieve this objective because we need to modify the existing MAC computation pattern and re-architect the hardware architecture to support new computing patterns. In the next section, we will elaborate how Tetris is designed for this purpose.

## III. Architecture of Tetris

### *A. Prerequisite*

As described in many previous literatures [5, 7], fixed point multiplication could be decomposed into a series of *shift-and-adds*, governed by the following Eq. (1):

$$A\times W = \sum_{b=0}^{B-1} 2^b \times (A\times W^b) \qquad (1)$$

If we have B length fixed point weight (W), the activation (A) is shifted b bits at each addition. The summation of these intermediate shifting segments denotes the final result. Similarly, we could extend this equation to multiple A/W pairs:

$$\sum_{i=0}^{N-1} A_i \times W_i = \sum_{b=0}^{B-1} 2^b \times \sum_{i=0}^{N-1} (A_i \times W_i^b) \qquad (2)$$

In the above Eq. (2), we firstly add all N number of As according to the *b*th bit of Ws, which could be either 0 or 1, and then perform the shift-and-accumulate on the final summation. As can be seen, the value of $W_i^b$ determines if the summation of $A_i$ is an ineffectual computation or vice versa. Tetris aims to replace these ineffectual bits with the essential bits of subsequent weights, as specified in the next subsection.

### *B. Weight Kneading Methodology*

The streamlined distribution of slacks could be utilized to shrink the ineffectual computations as minimum as possible. In Eq. (2), if the *b*th bit of weight $W_i$ is a slack, we seek to explore subsequent weights in the lane, i.e. $W_j$, and if its *b*th bit is an essential bit, the slack is replaced with $W_j^b$, and make the corresponding activation $A_j$ contribute to the current summation within the same cycle.

Figure 3 shows the general concept of this method. If we group 6 weights as a batch, conventionally it will take 6 cycles to accomplish 6 weight/activation MACs, because they are fed to PE one after another from on-chip eDRAM. Many previously proposed designs follow this paradigm [9, 15]. If we further interpret the weights and blind the zero bits, the slacks emerge at two orthogonal dimensions: (1) in the *intra*-weight dimension, i.e. $w_1$ and $w_6$, slacks demonstrates arbitrary distribution; (2) on the other hand, the slacks also show up at *inter*-weight dimension, i.e. $w_6$, who is an all-zero-bit (zero-value) weight, so it does not emerge at Figure 3(b). If we bubble up the essential bits taking the space previously occupied by the slacks, the computation cycles of 6 MACs will be decrease to 3 cycles as shown in Figure 3(c). This time we obtain $w'_1$, $w'_2$ and $w'_3$, but each one is combined with essential bits of $w_1 \sim w_6$. $w'_3$ has zero bits because we only take 6 weights as an example. If we allow more weights involved, these zero bits are likely to be filled by more essential bits, and we term this whole operation as "*Weight Kneading*" in this paper.

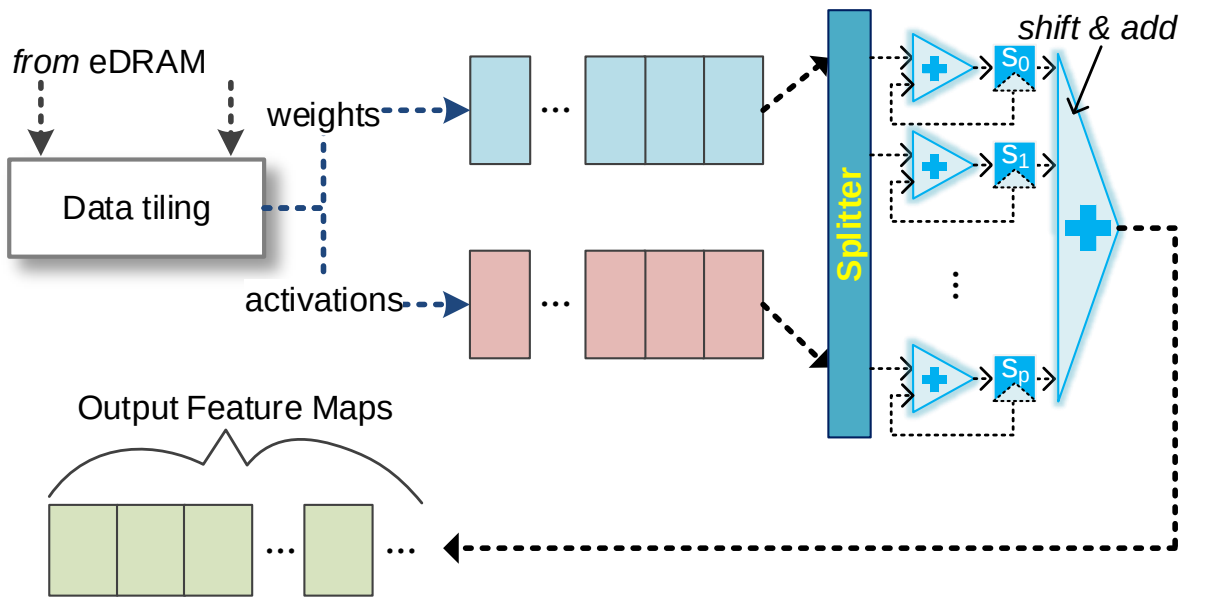

Figure 4 Split-and-accumulate (SAC). Differed from MAC, it does not aim to calculate the exact pair-wise multiplication; instead, it firstly splits a weight, and the activation is summed in the corresponding segment adders. Final shift-and-add for a partial sum is performed at last, right after certain amount of pairs accomplish processing.

Obviously, the benefits of weight kneading are, on one side, it automatically eliminate the impact of zero values without introducing extra operations like data preprocessing before feeding them into PE. As a small slice of total weights, the accelerator does not need to specifically deal with zero values, so the design complexity has the opportunity to be optimized. On the other side, zero bits are also replaced by essential bits after kneading. It avoids the impact of slacks at two dimensions at the same time. However, kneaded weights indicate the current summation is a combination of multiple corresponding activations instead of one activation alone, so the $A_i$ in Eq. (2) must be able to reference a set of activations according to the *b*th bit of $w'_i$. For example, if we configure 6 kneaded weights, then 6 corresponding activations must be reachable even if each kneaded weight may not need all of them, as Figure 3(c) shows: $w'_1$ only needs $A_1$, $A_2$ and $A_4$. The number of weights for kneading is a design parameter, termed as "*Kneading Stride* (KS)" and we will evaluate KS impact on the inference efficiency in our evaluation section.

### C. *From MAC to SAC*

To support such highly efficient computation, it is non-trivial to explore the necessity of the accelerator architecture that may be distinct with conventional ones aiming at serving classic MACs. Since we use the equivalent shift-and-add as the alternative to get one partial sum, each shifting segment is known as the combination of multiple contributive activations so the partial sum is not exactly the sum of a series of A/W multiplications. It is hence not necessary to shift *b* bits right after interpreting one kneaded weight w'. The final shift-and-add for *b* bits could be issued after pre-set number of kneaded weights or even after all the kneaded weights outright in the lane. This induces a new computation pattern that is different from the conventional MAC and canonical bit serialization schemes [5, 7], and we term it as "SAC" in this paper:

#### *1) Definition*

SAC denotes "**S**plit-and-**Ac**cumulate". A SAC operation firstly splits the kneaded weights, references the essential activations and finally accumulates each activation to the certain segment registers. Figure 4 shows the procedure of SAC for each A/W pair. Note that the figure shows *pair-wise* SAC as an example to demonstrate its concept, whereas in practical use, we resort to kneaded-weight SAC.

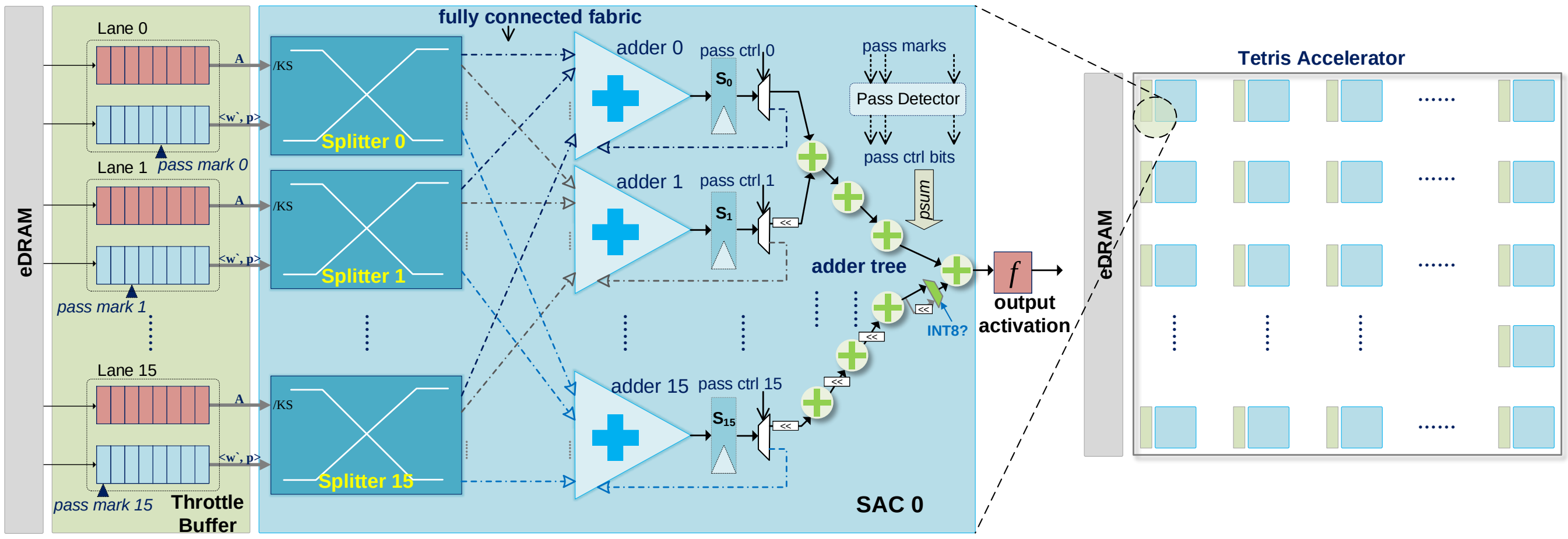

Figure 5 Tetris accelerator architecture. Each SAC unit is comprised of 16 parallel organized splitters accepting a series of kneaded weight and activation batches denoted by the kneading stride (KS) parameter, and same number of segment adders as well as a rear adder tree are responsible for calcualting the final partial sum.

SAC instantiates a “splitter” according to the bit length of weights: if we use fixed point 16 for each weight, we need 16 segment registers (p=16 in the figure) together with 16 adders with two input ports each. Typically, the splitter is responsible for dispersing each segment to its corresponding adder as governed by Eq. (2); for example, if the $2^{nd}$ bit of a weight is an essential bit, activation is delivered to S1 in the figure. Same operation applies to other bits. After “splitting” this weight, subsequent weights are handled in the same way, so each segment register is accumulated with new activation if its associate weight has an essential bit. After dealing with the lane outright, subsequent adder tree performs shift-and-add, once and for all, to obtain the final partial sum. Different from MAC, SAC does not perform any shift trying to obtain intermediate pair-wise multiplication value. The reason is that in real CNN model, output feature map only accounts for the “final” partial sum, that is, the summation of all channels of a filter and its corresponding input feature map. As for the intermediate partial sum, the value is not that useful so it is totally superfluous to waste time and energy procuring useless intermediate pair-wise partial sum as conventional MAC does. SAC computation pattern lies on this fact and move all shifting operations to the rear adder tree, and off the critical path of the multiplication

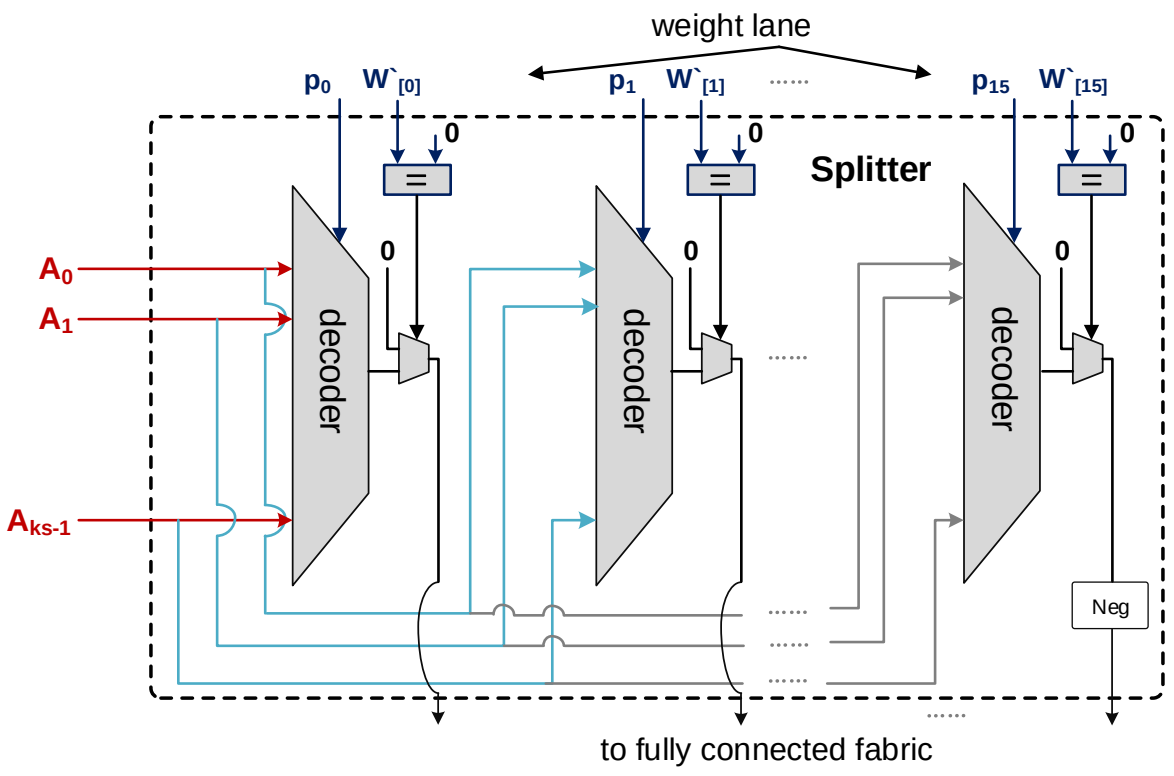

Figure 6 Microarchitecture of the splitter.

thereof. Especially when using kneaded weights, the benefits are more obvious compared with MAC.

### 2) *Tetris Accelerator Design*

In practical accelerator design, pair-wise SAC is not competent to boost the inference efficiency because it does not distinguish ineffectual computations, so it would still dispense activation to the segment even if the current bit of th e weight is zero. Therefore, we implement our Tetris accelerator that leverages *kneaded-weight SAC* to only distribute contributive activation to the segment adder for accumulation. Figure 5 shows the architecture of Tetris. Each SAC unit accepts addable A/W lane. Specifically, the accelerator consists of symmetric SAC units, connected with the throttle buffer accepting kneaded weights from the on-chip eDRAM. Each unit contains 16 splitters comprising a splitter array, if we use fixed point 16 weights. Each splitter is able to reference multiple impending activations and one kneaded weight, according to the parameter KS. There are 16 output data paths for activation to reach its target segment register for each splitter, so it forms a fully connected fabric between the splitter array and segment adders. For each segment adder, it receives activations from all 16 splitter for adding, as well as the values from local segment register. The intermediate segment sum is stored in S0~S15 register, and once all addable channel lanes are accomplished, “pass control signals” inform the multiplexer to pass each segment values to the rear adder tree. The last level of adder tree generates the final partial sum and passes it through to the output non-linear activation function and pooling. In the throttle buffer unit shown in the figure, the pass mark denotes the end of the addable A/W pairs, which will be notified to the pass detector in each SAC unit. If all pass marks reach the end, the adder tree is valid for the final summation of output feature map.

Since we use KS as a parameter to control the number of weights kneaded, different lanes will have different number of kneaded weights, so the pass marks, for most of the time, are not synchronized and may reside at any location of the throttle buffer. If new addable A/W pairs are filled into the buffer, the pass mark is moved backwards, so it does not impact the calculation of each segment partial sum.

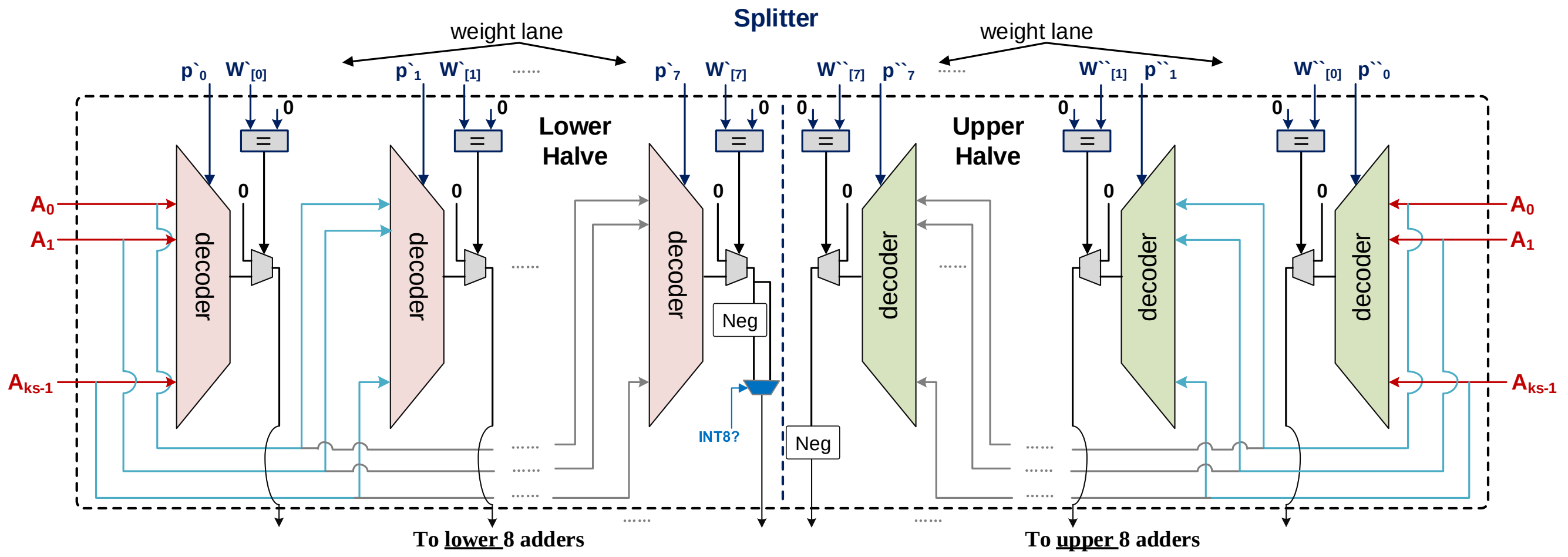

Figure 7 Modification of the splitter for accelerating int8 inference. Differed from fp16 mode, the splitter is halved into two parts with each handling an 8-bit width kneaded weight.

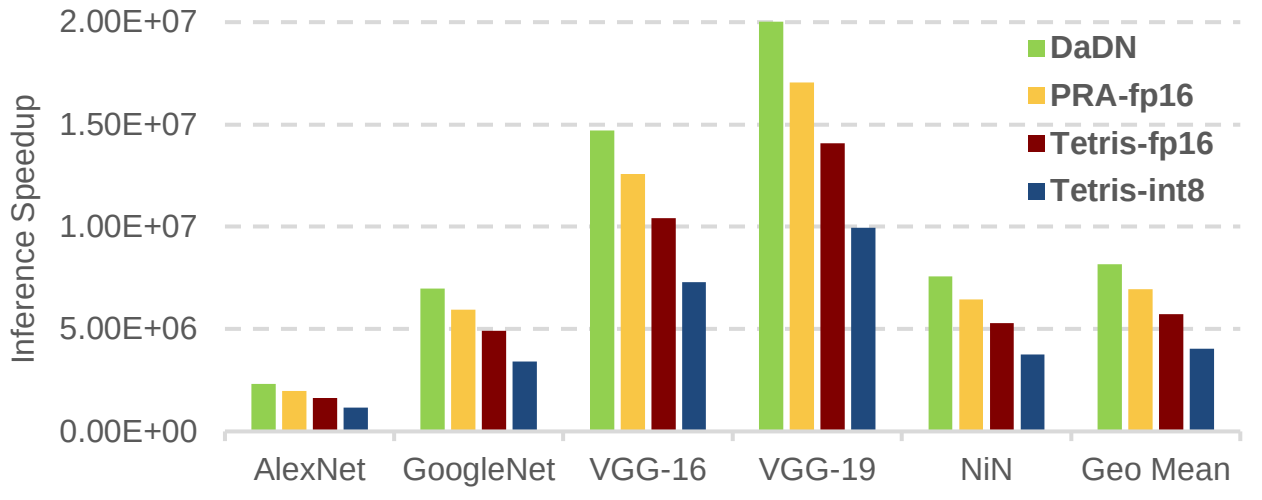

Figure 8 Performance comparison. We use absolute inference time consumed as the representative. Lower is better.

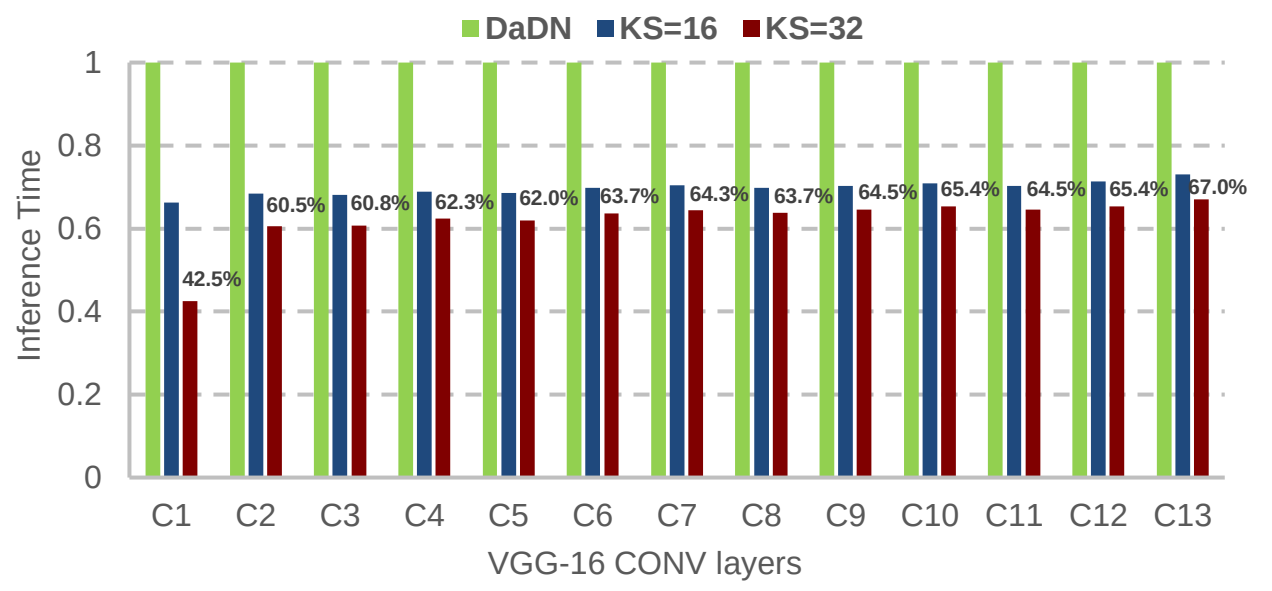

Figure 9 Speedup analysis of each Conv layer (normalized to DaDN) under two KS configurations of Tetris-fp16.

The splitter microarchitecture is shown in Figure 6. Weight kneading technique requires the splitter could accept a set of activations for one kneaded weight, and each essential bit must be recognizable to indicate the relevant activation value within the KS range. Tetris uses <w`$_i$, p> in the splitter to represent the activation associated with a particular essential bit. The bit-length of p is a proxy of KS, i.e. 4 bits p refers to a kneading stride of 16 weights. It allocates one comparator to determine if the input bit of a knead weight is zero, because it is possible that some bit locations are also ineffectual even after kneading, i.e. w’3 in Figure 3(c), depending on the KS. If it is a slack, the multiplexer after comparator will output zero to fully connected fabric. Otherwise, it will decode p and output corresponding activation among $A_0$~$A_{ks-1}$.

Although weight kneading scheme requires the splitter must be able to obtain any value within KS and transmit it to the segment adder through fully connected fabric, it does not mean the volume of the eDRAM must be also expanded. The splitter only needs to fetch the target activation in the throttle buffer when necessary, and the activations does not need to be stored for multiple times. The newly introduced p will consume a fraction of space but the vocation of p is only decoding target activation, and is only composed of several bits, i.e. 4 bits for 16 activations, so it will not introduce severe area and power overhead in the accelerator.

### 3) *8-bit Quantization Acceleration*

Tetris is capable of precision tunable acceleration, which is another benefit of SAC. Some prior work has proved the accuracy and precision of the DCNN model demonstrate a tradeoff, and a graceful accuracy degradation is acceptable when the precision is tuned and decreased a little bit for each layer [7]. Tetris does not need to modify its intrinsic architecture to fulfill this purpose, because the contributive activations are bit dependent of kneaded weights. If we shrink the length of weights from fixed point 16 to arbitrary lengths, i.e. 8, 9 or even 4 bits, SAC operates itself in the same way as fp16, with only one distinction that not all the adders after fully connected fabric are active. If we use 4-bit weight, only adder0 ~ adder3 remain activated because they accumulate segment 0 to 3 according to the weight bit length.

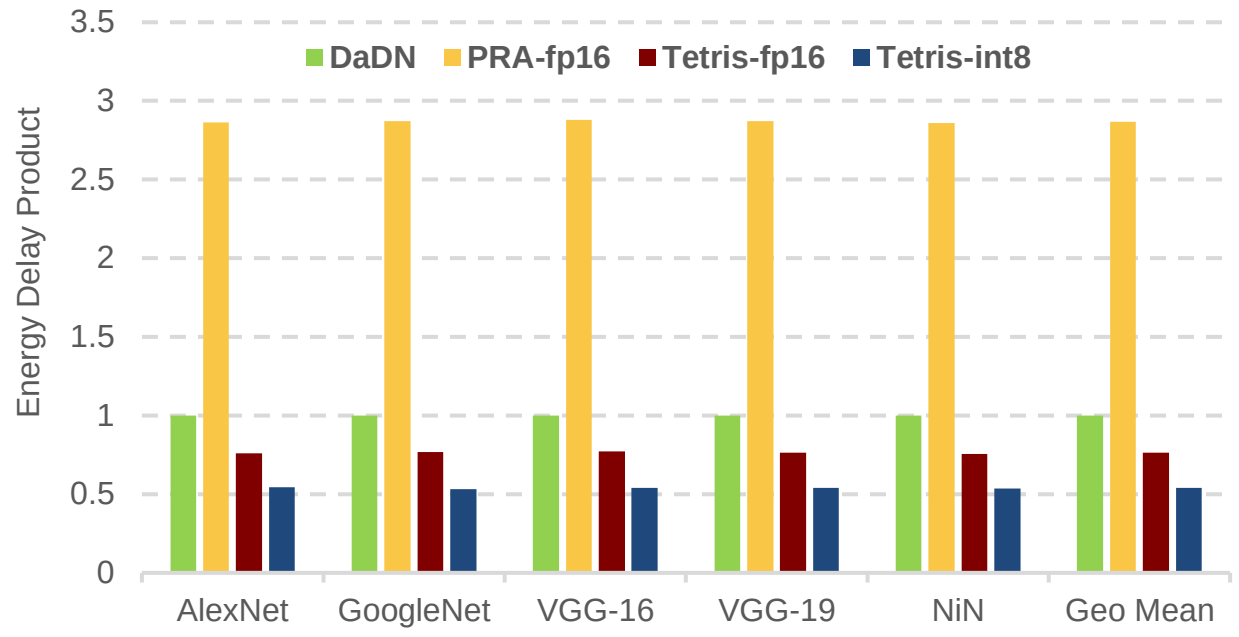

Figure 10 Energy efficiency comparsion, normalized to DaDN.

In recent years, 8-bit integer quantization is proved to be amenable to high throughput inference and even training. Many deep learning framework like Tensorflow and Caffe tends to incorporate INT8 arithmetic as the basic operation in stochastic gradient descent calculation. It is also leveraged by many types of GPUs and software programmable engines like TensorRT [16] to accelerate inference of DCNN models. As the unique feature of Tetris, it could provide a doubled inference efficiency compared with fp16 mode which makes Tetris promising in deploying DCNN models in lightweight devices. By simply configuring the splitter array and segment adders, Tetris could be configured to INT8 mode as Figure 7 shows. The fundamental architecture of SAC unit remains nearly unchanged except for the splitter. The splitter is firstly divided into two halves, with each halve accepting one kneaded weight. It turns out that the upper 8 adders will deal with the upper halve activations, similar for the lower 8 adders. Besides, only the last level of the rear adder tree needs to be configured to distinguish the two modes. By this manipulation, the segment adders and rear adder tree are both sufficiently utilized without idle components. Since each splitter takes two kneaded weights as input, so the throughput of the SAC unit is doubled with the same KS setting, which means under the same volume of activation inputs, the inference efficiency would also be twice in theory compared with fp16 mode.

## IV. Evaluation

In this section, we evaluate the proposed Tetris accelerator. The DCNN model definitions and their pre-trained synaptic weights are obtained from Caffe Model Zoo [17].We quantize the initial floating point 32 weights into fixed point 16 and integer 8 precision. Then, we fine-tune the weights using framework Caffe [18] to maintain the *Top-1* classification accuracy. The obtained fp16 and int8 weights are used for weight kneading in evaluating inference efficiency of various DCNN models.

We use two baselines: the 1st one is DaDianNao [9], the de facto design to report relative performance of diverse DCNN accelerators; the 2nd one is a state-of-the-art bit-serial implementation [5] (PRA), it is also designed for computing essential bits of the activations, and we enroll its fp16 design on weights for fair comparison. We implement Tetris with two configurable modes, fp16 and int8, as mentioned in Section III.

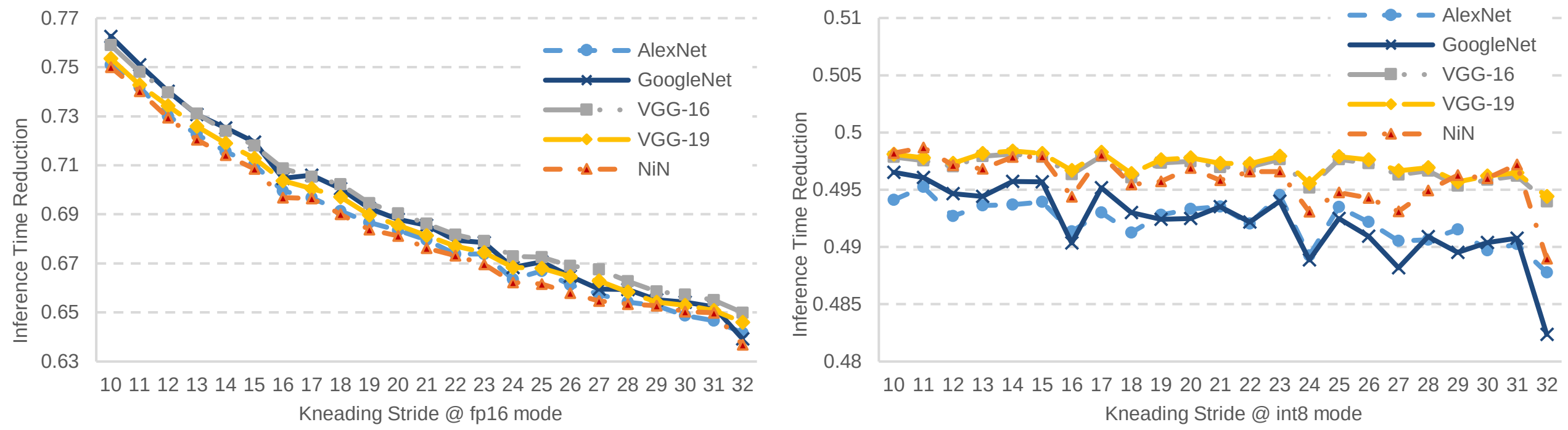

Figure 11 Performance under different kneading stride (KS) configurations for fp16 (upper) and int8 (lower) mode. We use $T_{ks}/T_{base}$ in plotting this figure, where $T_{ks}$ is the time consumed after kneading ks number of weights, while $T_{base}$ is the time that weight kneading is not applied.

Table 2 Area overhead comparison. We evaluate total area of Tetris as well as the baselines, together with area breakdown for 1 PE of Tetris.

| **Area (16 PEs) in mm$^2$** | | **Area Breakdown for 1 PE of Tetris (mm$^2$)** | | | | | | |
|---|---|---|---|---|---|---|---|---|
| DaDN | 79.36 | **ITEM** | I/O RAMs | Throttle Buffer | Splitter Array | Activation Function | Segment Adders | Rear Adder Tree |
| PRA-fp16 | 153.65 | **Size** | 20KB/PE | 5KB | 16x 16SACs | ReLu | 16x 16SACs | 1x 16SACs |
| Tetris-fp16 | 89.76 | **Area** | 3.828 | 0.957 | 0.544 | 0.143 | 0.129 | 0.008 |
| **Overhead** | 1.131048 | **Percentage** | 68.24% | 17.06% | 9.70% | 2.55% | 2.30% | 0.14% |

We employ Vivado HLS (v2016.2) to conduct C simulation and C/RTL hybrid simulation to extract the inference time of various DCNN models. For area evaluation, we compile our design using Synopsis Design Compiler [19] with TSMC 65nm technology library. For power/energy measurement, we use PrimeTime tool [20] after HLS for the intrinsic components of Tetris and other baselines. We use 16 PEs for Tetris and the baselines and each PE is clocked at 125MHz. This frequency setting is referenced in HLS simulation for Xilinx Z7020 FPGA, for which fp16 multiplications could be accomplished within one cycle. The practical frequency for our proposed Tetris is possible to be tuned higher because the replacement of multipliers with multi-operand adders provides abundant timing intervals which could be utilized to boost the inference frequency, but in this paper we keep the frequency constant for fair evaluation. For the only design parameter KS, we choose 16 weights to be kneaded, in other words, the input splitter could reference 16 activations at one time according to the input kneaded weight. This design parameter is also evaluated to explore the sensitivity of Tetris to inference speedup.

## *A. Performance*

Figure 8 shows the inference speedup of baselines as well as two mode implementations of Tetris, using real world execution cycles in Vivado HLS simulation as the representative. Here, we observe that by kneading weights Tetris could achieve 1.30x speedup for fp16 and 1.50x for its int8 mode over DaDN on average. Comparatively, the other baseline PRA could achieve nearly 1.15x speedup. The benefits comes from parallel referencing multiple activations according to the kneaded weights in the synaptic lane. With deploying split-and-accumulate (SAC) instead of MAC in Tetris, weight kneading is properly supported at the hardware level. Partial sum is not computed at pair-wise pattern, and Tetris eliminate the impact of ineffectual computations that is ignored by DaDN. Dealing with activation data in batches significantly improves the throughput of inference. PRA-16fp is a bit serial computing scheme and it must traverse the entire weight to probe essential bits, and leverage large shifters at multiple stage shifting to accumulate the partial sum. The whole operation cannot be accomplished within one cycle. Since it also accounts for the essential bits, inference is also accelerated but only at small magnitude, while Tetris does not incur this problem. Combined with specialized int8 acceleration mode, Tetris provides nearly doubled inference speedup. Figure 9 shows per-layer speedup analysis of VGG-16 as a case st udy.

## *B. Energy Efficiency*

Furthermore, we compare the energy efficiency of Tetris and the baselines. This time we use energy delay product as the proxy. After the whole DCNN model finishes inference for a single image, we trace the total cycles consumed together with power consumption after synthesis and calculate final EDP values. To facilitate direct comparisons across all DCNN models and the baselines, we normalize the EDP results to DaDN reported in Figure 10. The average efficiency improvement of Tetris is 1.24x for fp16 and 1.46x for int8 compared with DaDN. The benefit of performance enhancement does not bring with the cost on energy for Tetris. Tetris does not introduce complex hardware architecture to fulfill SAC and on the contrary, it simplifies the multiplier to pure adders. But from our evaluation, it takes 1.08x average power increase due to multiple pre-adding splitters and multi-input adder trees, but the small power increase is overshadowed by the abundant inference speedup. For bit-serial baseline, it must enroll 16x more weight buffers to compensate the throughput loss compared with DaDN, so power consumption increases enormously to 3.37x, and degrades the power efficiency to 2.87x compared with DaDN. Tetris outperforms PRA as 3.76x and 5.33x for fp16 and int8 mode respectively.

### C. Sensitivity Analysis

Kneading Stride, as the major design parameter, affects the runtime inference speedup since the performance primarily depends on the amount of weight/activation pairs computed. Intuitively, the more weights kneaded, the more speedup obtained. That is because more kneaded weights will lead to more opportunities to fill up zero slacks that further increases the effective computations and throughput. Besides, more kneaded weights reduces the operations of computing partial sum, so it shortens the inference time that further contributes to more energy savings. Our evaluation proves this notion as shown in Figure 11. We scale the KS from small (10 weights) to large (32 weights) and evaluate the inference time reduction at each step. The total time is saved to 75.1% for AlexNet at KS=10, and further reduced to 64.2% at KS=32. For int8 mode, the data are ranged from 49.4% to 48.8%. Other DCNN models exhibit similar behaviors at fp16 mode. However, we cannot naively set KS as large as possible, because the interleaved essential bits also leads to wider decoders in the splitter. It must be capab le to reference more activations at a time by relying on more bits of <p> in Figure 6. To acquire a balanced design complexity and inference speedup, we choose KS=16 in conducting previous evaluations. However, this design parameter could be dynamically configured at different scenarios with different speedup and power constraints.

### D. Area Overhead

Table 2 lists the area of Tetris and two baselines. The overall area overhead is 1.13x compared with DaDN, but Tetris has a relatively smaller area compared to PRA. PRA suffers from large weight FIFOs, because the bit serialization cannot match the throughput of bit parallel schemes like DaDN and Tetris, and large buffers must be introduced to provide more bit level operation simultaneously [5]. The overhead of 16 PEs could reach 1.93x over DaDN. For the area breakdown, the major contributor for Tetris is I/O activation/weight RAMs allocated per PE and throttle buffer (68.24%, 17.06%). For functional components, splitter array (9.7%) dominates chip area. Segment adders and read adder tree do not occupy a large space with $0.1293mm^2$ and $0.008mm^2$ each.

## V. Conclusion

In this paper, we propose a novel computing paradigm as well as the Tetris accelerator that leverages weight kneading and SACs targeting efficient inference of modern DCNN models. Differed from preconceived design philosophy, Tetris does resort to MAC operation for obtaining exact partial sums with the cost of tedious multiplying and shifting, and most critically oblivious to the zero bit slacks that leads to ineffectual computation. Weight kneading technique assures that only the essential bits are involved in DCNN inference and SAC computing pattern is design to fulfill this purpose at the architectural level. We believe that the techniques proposed in this paper will motivate a reconsideration of DCNN accelerator design by applying the same concept over other designs or general purpose computing engines like GPUs in the future.